\documentclass[letterpaper]{article} 
\usepackage[preprint]{aaai2027} 
\usepackage[hyphens]{url} 
\usepackage{graphicx} 
\usepackage{natbib} 
\usepackage{caption} 
\usepackage{amsmath,amssymb,amsfonts}
\usepackage{booktabs}
\usepackage{makecell}
\usepackage{multirow}
\DeclareMathOperator*{\argmax}{argmax}

\title{Beyond Block Boundaries: Multi-Block Editing for Diffusion Large Language Models}
\author{
Xingyu Mou$^{1}$ \quad
Zijin Huang$^{1}$ \quad
Tianze Zhang$^{1}$ \quad
Yuxin Ma$^{1}$ \\
Lanning Wei$^{1}$ \quad
Zengfeng Huang$^{2,3}$ \quad
Da Zheng$^{1}$ \quad
Lun Du$^{1}$\corresponding
}

\affiliations{
$^{1}$Ant Group \qquad
$^{2}$Fudan University \qquad
$^{3}$Shanghai Innovation Institute\\
\texttt{dulun2834@gmail.com}
}
\begin{document}
\maketitle

\begin{abstract}
Block diffusion has emerged as the dominant paradigm for scaling discrete diffusion language models (dLLMs), because decoding text in fixed-size blocks preserves parallel generation within each block while keeping the quadratic attention cost tractable. However, this efficiency comes with a structural limitation: tokens near the end of a block are generated without access to future cross-block context, and once a block is finalized, its uncertain predictions become irreversible context for all subsequent blocks. This creates a \emph{block boundary problem}, in which sensitivity to missing future context grows toward block boundaries and early mistakes propagate throughout later generation. Comparing predictions with and without later-block context
reveals sharply increasing boundary sensitivity: on AIME~2025, the mean self-containedness divergence (SCD) in the last quarter of a block is 61.3$\times$ that in the first quarter. To address this issue, we propose \textbf{Multi-Block Editing (MBE)}, which edits decoded tokens based on cross-block context. MBE first introduces a training-free decoding algorithm that reopens a full-attention window over selected blocks. Given the mismatched attention mechanism between block diffusion training and MBE inference, MBE further introduces Multi-Block Edit SFT, which equips the model with bidirectional attention masks and progressively expands the editing span. We also extend SGLang with a multi-shape CUDA Graph pool and fine-grained KV-cache control to make variable-length editing passes efficient in practice. 
Experiments on LLaDA2.1-Mini across 12 benchmarks
demonstrate broad and consistent gains. Without any parameter
updates, Training-Free MBE consistently improves or matches standard
decoding on every benchmark. The full MBE framework raises
the 12-benchmark average from 61.45 to 64.24, with gains of
up to 13.33 points on AIME~2025.
MBE retains 87.3--96.7\% of standard-decoding end-to-end throughput across four datasets.
\end{abstract}

\section{Introduction}

Discrete diffusion large language models (dLLMs), such as
LLaDA~\citep{nie2025large}, Dream~\citep{ye2025dream}, and
Seed Diffusion~\citep{song2025seed}, generate text by
denoising multiple tokens in parallel, offering substantially
higher throughput than autoregressive (AR) models. Block
diffusion~\citep{arriola2025block} further improves
scalability by decoding fixed-size blocks left-to-right while
denoising tokens within each block in parallel, thereby
limiting the active diffusion span and keeping attention costs
manageable. This design has become a practical approach for
scaling dLLMs to large model sizes: recent 100B-scale
systems, including LLaDA2.0~\citep{bie2025llada2} and
LLaDA2.1~\citep{bie2026llada2}, adopt it to balance
inference efficiency against the quality trade-offs of
blockwise decoding.

\begin{figure}[t]
  \centering
  \includegraphics[width=\columnwidth]{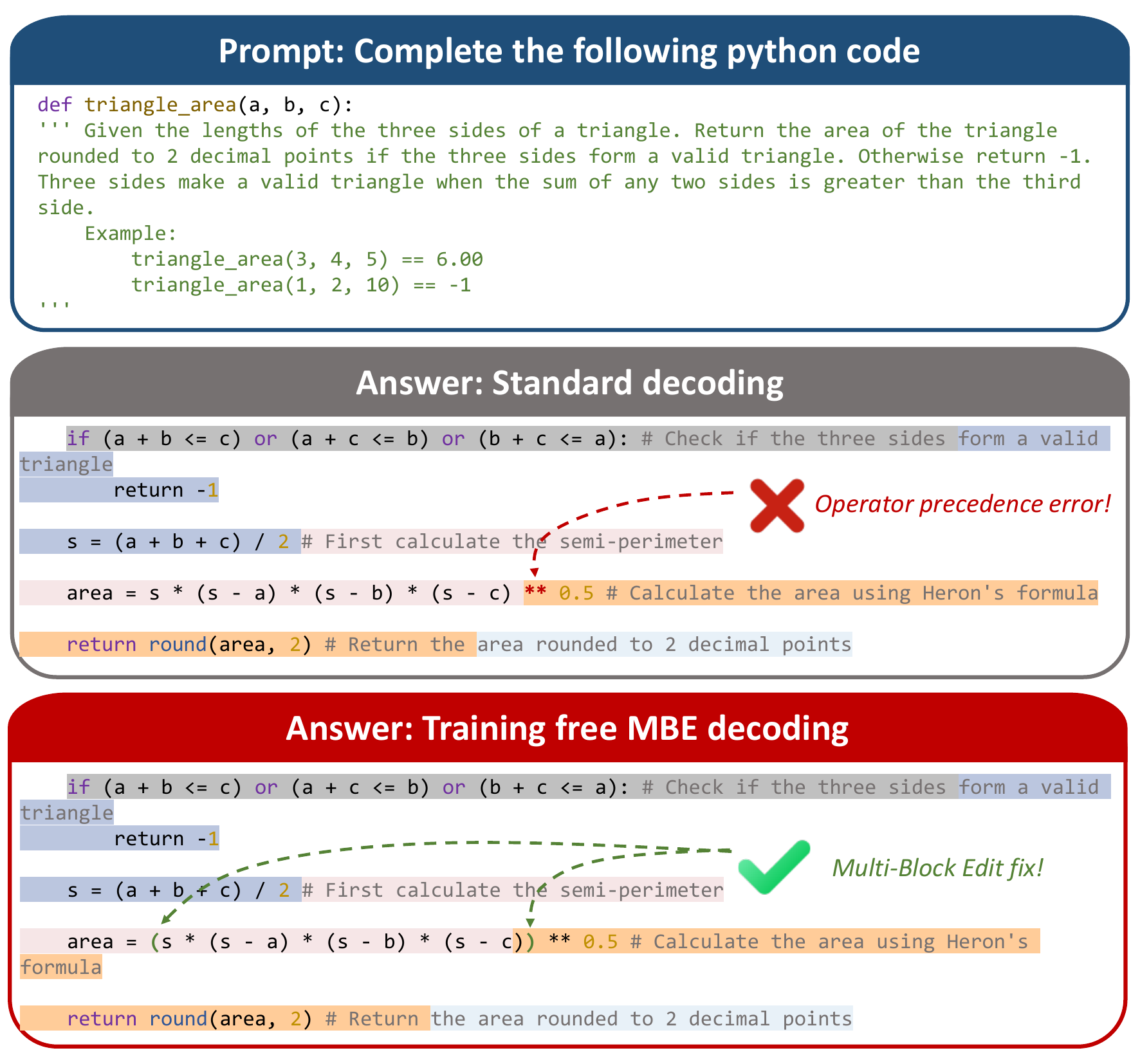}
  \caption{A boundary-induced HumanEval+ error corrected by MBE. Standard decoding commits an operator-precedence error before the exponent is visible. Once the next block is available, MBE reopens the completed code and inserts the required parentheses.}
  \label{fig:leadin}
\end{figure}

This efficiency, however, introduces a structural limitation. Although all tokens in a block are decoded without access to subsequent blocks, the effect is strongest near the block end, where less within-block right context is available. More
importantly, block diffusion follows a decode-and-commit paradigm: once finalized, a block becomes fixed context for all subsequent generation and cannot be revised when new context arrives. Figure~\ref{fig:leadin} illustrates this issue
on HumanEval+: Standard Decoding commits an expression without the required parentheses before the exponent appears in the next block, whereas reopening the completed code corrects the resulting operator-precedence error. We refer to
this boundary-localized, irreversible degradation as the \emph{block boundary problem}. Position-wise analyses of future-context sensitivity and predictive uncertainty confirm that both effects concentrate near the block end (Section~\ref{sec:motivation}), motivating revision after commitment.

A natural remedy is to revisit finalized blocks after later blocks have been decoded, allowing earlier predictions to be revised using richer cross-block context. Realizing this idea, however, poses three challenges. \textbf{(i)~Obtaining cross-block context.}
Standard block diffusion restricts each forward pass to a single block, whereas correction requires joint passes over multiple decoded blocks, changing the model's attention
scope. \textbf{(ii)~Efficiency and simplicity.}
Revision must preserve the parallel-decoding advantage of dLLMs while adding limited computation and avoiding complicated token-level control. \textbf{(iii)~Training--inference mismatch.} Pretrained block diffusion models are optimized for intra-block denoising rather than multi-block revision, so wider editing windows may be ineffective without explicit alignment.

We address these challenges with
\textbf{Multi-Block Editing (MBE)}. Training-Free MBE reopens two-block full-attention windows to revise committed tokens using generated later-block context. \textbf{Multi-Block Edit SFT} supports wider editing windows through a future-aware cross-block attention mask and a span
curriculum, while tail-entropy selection determines where each editing window begins. A multi-shape CUDA Graph pool and fine-grained KV-cache control enable efficient variable-length editing in SGLang. Our contributions are:

\begin{itemize}
  \item We identify the \emph{block boundary problem}, characterize its positional concentration, and theoretically establish when later-block context can improve committed predictions.

  \item We propose MBE as a unified cross-block decoding and post-training framework. Training-Free MBE revises committed blocks using generated later-block context, while Multi-Block Edit SFT aligns the model with wider editing windows.

  \item We implement MBE efficiently in SGLang and validate it across 12 benchmarks, multiple block sizes, and LLaDA2.1 model variants. The full framework raises the   12-benchmark average from 61.45 to 64.24 while retaining   87.3--96.7\% of standard-decoding throughput.
\end{itemize}

\section{Preliminaries}

\subsection{Discrete diffusion language models}

A discrete diffusion language model generates $\mathbf{x}=(x_1,\ldots,x_L)$ by iteratively denoising a masked sequence. The forward process replaces clean tokens with \texttt{[MASK]} according to a noise schedule; the reverse model $p_\theta(\mathbf{x}_0\mid\mathbf{x}_t)$ predicts the clean sequence. Inference alternates model forwards with mask-to-token updates until no masks remain.

\subsection{Block diffusion}

LLaDA2.1~\citep{bie2026llada2} divides the output into $B$-token blocks, decodes blocks left-to-right, and denoises tokens within each block in parallel. Block $j$, spanning $[jB,(j{+}1)B)$, conditions on the prompt and completed prefix $\mathbf{x}_{0:jB}$. This bounds the active diffusion span to $B$ tokens; once committed, a block is not revisited by Standard Decoding.

\subsection{Configurable threshold decoding}
\label{sec:threshold_decoding}

LLaDA2.1 jointly performs Mask-to-Token (M2T) and Token-to-Token (T2T) updates. At step $t$, let $\hat x_i^{(t)}$ be the highest-probability prediction at position $i$, with confidence $c_i^{(t)}=p_\theta(\hat x_i^{(t)}\mid\mathbf{x}^{(t)})$. For the masked-position set $\mathcal M_t$, the two update sets are
\begin{equation}
\begin{aligned}
\Gamma_t&=\{i\in\mathcal M_t:c_i^{(t)}>\omega_{\rm mask}\},\\
\Delta_t&=\{i\notin\mathcal M_t:\hat x_i^{(t)}\ne x_i^{(t)},\ c_i^{(t)}>\omega_{\rm edit}\}.
\end{aligned}
\label{eq:threshold_updates}
\end{equation}
Positions in $\Gamma_t$ are unmasked to $\hat x_i^{(t)}$, while visible tokens in $\Delta_t$ are overwritten by $\hat x_i^{(t)}$; both updates use the same forward prediction. If $\Gamma_t$ is empty while masks remain, the highest-confidence masked position is force-unmasked, guaranteeing resolution within at most $B$ steps. LLaDA2.1 then performs up to 16 T2T-only post-edit passes, stopping early when no token changes, before permanently committing the block. MBE retains the same T2T acceptance rule but replaces this intra-block post-edit phase with refinement after later-block context becomes available.

\section{Empirical Observations}
\label{sec:motivation}

Standard block decoding commits a block before any later block has been generated. We first ask where those later blocks would most strongly change its predictions. VSB~\citep{wang2026vsb} averages token-level No-Future/Future-Aware divergences within a candidate block for boundary selection. We instead aggregate the same divergence by relative position $r$ within a fixed block:
\begin{equation}
\operatorname{SCD}(r)=\frac{1}{|\mathcal I_r|}
\sum_{x\in\mathcal I_r}
D_{\rm KL}\!\left(q^{\rm NF}_{r}(\cdot\mid x)\,\|\,q^{\rm FA}_{r}(\cdot\mid x)\right),
\label{eq:scd}
\end{equation}
where $q^{\rm NF}_{r}$ and $q^{\rm FA}_{r}$ are the predictive distributions before and after later context is exposed, and $\mathcal I_r$ contains the retained occurrences of position $r$. A larger SCD means that the prediction changes more after future context becomes available, indicating stronger dependence on information absent at commitment.

\begin{figure}[t]
  \centering
  \includegraphics[width=\columnwidth]{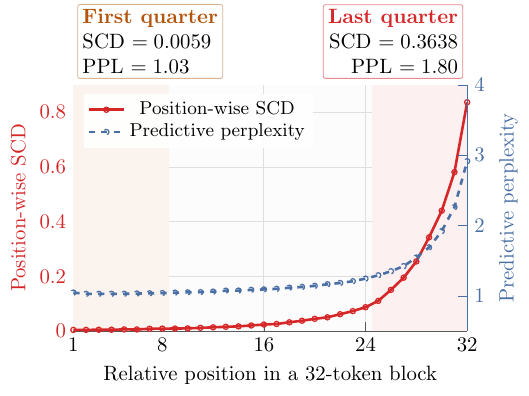}
  \caption{Position-wise SCD (red, left axis) and predictive perplexity (blue, right axis) on AIME~2025. Shading and labels compare first- and last-quarter means.}
  \label{fig:position_observations}
\end{figure}

Using LLaDA2.1-Mini with $B{=}32$, we collect several thousand decoded blocks and approximately $1.8{\times}10^5$ valid token-position observations on all AIME~2025 problems. SCD rises sharply toward the block boundary (Figure~\ref{fig:position_observations}), with the last-quarter mean approximately $61.3{\times}$ that of the first quarter.

This dependence also affects the block-local prediction itself. Because the relevant future context is unavailable during Standard Decoding, boundary positions must be resolved from less complete information. For the commit-time distribution $q_r$, we measure this uncertainty using
\begin{equation}
\operatorname{PPL}(r)=
\exp\!\left[-\sum_{v\in\mathcal V}q_r(v)\log q_r(v)\right].
\label{eq:ppl}
\end{equation}
For a uniform distribution over $n$ tokens, $\operatorname{PPL}(r)=n$; more generally it is the effective number of equally likely candidates. Under Standard Decoding, perplexity rises from 1.03 in the first quarter to 1.80 in the last and reaches 2.92 at the final position. The positions most sensitive to future context are therefore already the most uncertain when Standard Decoding commits them.

\section{Methodology}

\begin{figure*}[t]
  \centering
  \includegraphics[width=0.94\textwidth]{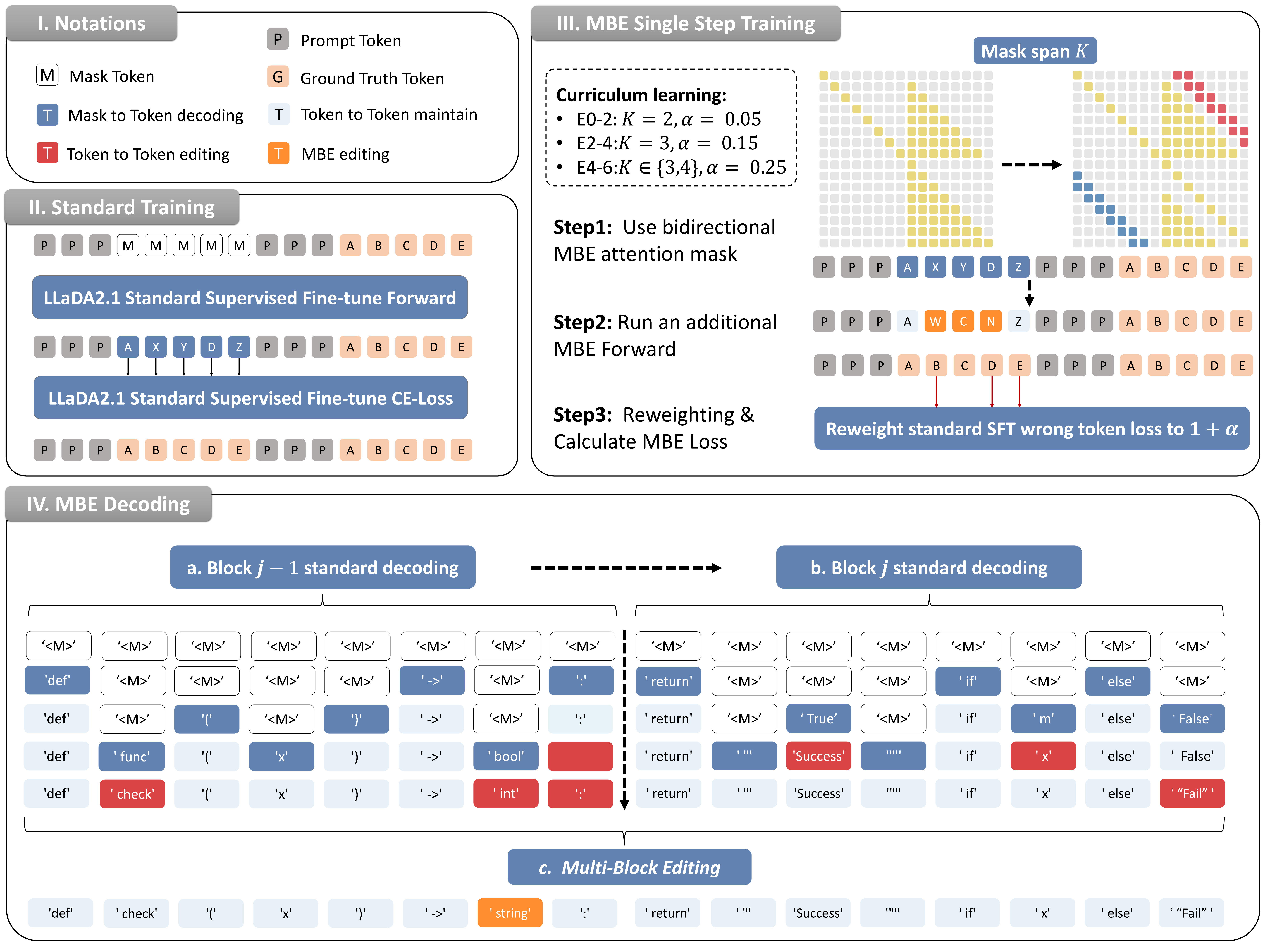}
  \caption{The MBE framework. Multi-Block Edit SFT (upper panels) augments standard SFT with a future-aware cross-block mask, curriculum, and correction-weighted loss. Training-Free MBE (lower panel) reopens completed tokens and applies T2T edits. Runtime support for variable forward shapes and cache updates is described in Section~\ref{sec:system}.}
  \label{fig:mbe_overview}
\end{figure*}

MBE combines cross-block decoding, attention-aligned post-training, and runtime support for variable-shape forwards and cache-safe revision (Figure~\ref{fig:mbe_overview}).

\subsection{Training-Free MBE Decoding}
\label{sec:mbe_inference}

Training-Free MBE reopens a short window after the current block converges, revising committed tokens with generated later-block context without changing model parameters.

\paragraph{Inference algorithm.}
Block $j$ spans $[jB,(j{+}1)B)$. After standard intra-block M2T/T2T denoising converges, MBE selects a historical start block $j^\star$ and, for $j>0$, opens
\begin{equation}
W_j=[j^\star B,(j{+}1)B).
\label{eq:window}
\end{equation}
Each pass uses full attention within $W_j$ and applies T2T edits only to its historical portion $[j^\star B,jB)$. MBE replaces the usual post-resolution intra-block edit phase with at most $N_{\rm MBE}$ cross-block passes, stopping when no token changes. Modified blocks are then recomputed under standard attention to refresh their KV cache.

\paragraph{Two-block window.}
Training-Free MBE fixes $j^\star=j{-}1$, exposing the next decoded block while limiting attention-pattern shift. Complete pseudocode is in the Supplementary Material.

\paragraph{Tail-entropy block selection.}
For wider windows, let $e_{i,r}=H(p_\theta(\cdot\mid\mathbf{x};iB+r))$ be the final standard-scope entropy at position $r$ of block $i$. If $\mathcal T_\rho(i)$ contains its $\lceil\rho B\rceil$ largest-entropy positions, define
\begin{equation}
\mathcal S_\rho(i)=\frac{1}{|\mathcal T_\rho(i)|}
\sum_{r\in\mathcal T_\rho(i)}
e_{i,r},
\label{eq:tail_entropy}
\end{equation}
where $H(p)=-\sum_vp_v\log p_v$. The upper-tail mean retains localized uncertainty without reducing a block to one maximum-entropy token. Among $\mathcal C_j=\{i\mid\max(0,j-K_{\max}+1)\le i<j\}$, MBE selects
\begin{equation}
j^\star=\argmax_{i\in\mathcal C_j}\mathcal S_\rho(i).
\label{eq:block_selection}
\end{equation}
The selected block determines how far revision reaches into history. The resulting window remains contiguous through the current block, so all intervening later context is preserved. We use $K_{\max}{=}4$; a monotonic deque maintains the maximum score with amortized $O(1)$ selection overhead.

\paragraph{Computational cost.}
An MBE pass processes at most $KB$ tokens; bounded passes and early exit limit the overhead measured in Section~\ref{sec:efficiency}.

\subsection{Training: Multi-Block Edit SFT}
\label{sec:mbe_sft}

Training-Free MBE already yields consistent gains with a two-block editing window, but directly expanding the window beyond $K{=}2$ introduces attention patterns that the pretrained model has not observed. We therefore design Multi-Block Edit SFT to align the model with the joint attention used by wider MBE windows and extend the editing range to $K_{\max}{=}4$.

\paragraph{Cross-block attention mask.}
Standard block SFT contains noisy-to-noisy (NN), noisy-to-clean-prefix (NC), and clean-to-clean (CC) links. We add noisy-sees-forward (NF) links from noisy block $i$ to clean blocks $i{+}1,\ldots,i{+}K{-}1$, simulating revision with later context, plus corresponding clean-sees-noisy (CN) links. To prevent an indirect path from a noisy block to its clean target, implicated CC links are removed. Prompt tokens remain a separate category; the full construction is in the Supplementary Material.

\paragraph{Curriculum and correction bonus.}
The curriculum progressively expands the attention span $K$ while increasing a correction bonus $\alpha$, so the model learns short-range revision before receiving wider cross-block patterns. Standard M2T/T2T objectives remain active. At final denoising, the loss at position $u$ is multiplied by $1+\alpha\,\mathbf 1[\widehat y_u^{\rm init}\ne y_u]$, concentrating additional supervision on initially incorrect tokens that later context can correct. Exact phases and training settings are reported in Section~\ref{sec:experiments} and the Supplementary Material.

\subsection{Infrastructure Optimization}
\label{sec:system}

MBE breaks two assumptions in the standard LLaDA2.1 decoding path in SGLang~\citep{zheng2024sglang}: forward passes within a request have a fixed shape, and every pass updates the KV cache in the same way. An MBE request instead alternates between $B$-token standard forwards and joint editing forwards of up to $KB$ tokens. Moreover, the joint pass uses a wider attention scope, so its KV states must not replace the standard-scope states consumed by subsequent block decoding. Efficient execution therefore requires both multi-shape graph capture and explicit cache-update control.

\paragraph{Multi-shape CUDA Graph pool.}
We extend the CUDA Graph runner with a graph pool indexed by batch size and raw forward length. For each supported batch size, graphs for the single-block and joint-window shapes are captured at startup and share a statically allocated maximum-shape input buffer; shorter inputs use tensor views into that buffer. FlashInfer~\citep{ye2025flashinfer} attention metadata is indexed by the same pair, ensuring that each shape receives the correct attention plan. Runtime dispatch then selects the graph matching the current forward length instead of falling back to eager execution for editing passes.

\paragraph{Cache control.}
The joint editing pass reads cached history but sets a cache-control flag that suppresses KV writes. Otherwise, full-attention representations from the editing window would overwrite standard-scope entries and leave later blocks conditioning on inconsistent states. MBE records which historical blocks actually changed and, after editing converges, recomputes only those blocks under standard attention. This targeted refresh restores cache consistency without rebuilding the full window.

\section{Theoretical Analysis}
\label{sec:theory}

\noindent\textbf{Commit-time information constraint.}
Consider block $i$, with target tokens $Y_i=\mathbf{x}_{iB:(i+1)B}$. Let $C_i$ collect the prompt, committed prefix, and within-block decoding states available when Standard Decoding commits this block. We distinguish two later-block contexts: $F_i^\star$ contains the ground-truth tokens in the editing window, whereas $\widetilde F_i$ contains the corresponding tokens generated after block $i$ is committed. In general,
\begin{equation}
  \widetilde F_i=G_\theta(C_i,\xi),
  \label{eq:generated_future}
\end{equation}
where $\xi$ represents decoding randomness and is omitted under deterministic decoding. Standard and MBE predictions can then be written as
\begin{equation}
  \widehat Y_i^{\rm Std}=A_{\rm Std}(C_i),
  \qquad
  \widehat Y_i^{\rm MBE}=A_{\rm Edit}(C_i,\widetilde F_i).
  \label{eq:commit_constraint}
\end{equation}
Every M2T, T2T, and post-edit update in Standard Decoding is conditioned only on $C_i$. Additional intra-block updates may change $A_{\rm Std}$, but cannot introduce the future semantic evidence represented by $F_i^\star$. MBE instead generates a proxy $\widetilde F_i$ for this unavailable context and revisits the committed block using that proxy.

\noindent\textbf{Theorem 1 (Oracle value of ground-truth future context).}
For a loss $\ell$, let $\mathcal R_\ell^*(Z):=\inf_a\mathbb E[\ell(Y_i,a(Z))]$ denote the Bayes-optimal block risk under information $Z$. Augmenting the commit-time context with ground-truth later blocks satisfies
\begin{equation}
  \mathcal R_\ell^*(C_i,F_i^\star)
  \leq \mathcal R_\ell^*(C_i).
  \label{eq:value_future_context}
\end{equation}
Under log loss, the reduction in Bayes risk is exactly
\begin{equation}
  \mathcal R_{\log}^*(C_i)
  -\mathcal R_{\log}^*(C_i,F_i^\star)
  =I(Y_i;F_i^\star\mid C_i)\geq 0.
  \label{eq:log_value}
\end{equation}
Thus, $I(Y_i;F_i^\star\mid C_i)$ quantifies the oracle value of future-block information unavailable at commitment. The complete proof is in the Supplementary Material.

\noindent\textbf{Practical gain with predicted future context.}
MBE observes $\widetilde F_i$, not $F_i^\star$. Let $q_{\rm Std}(y\mid c)$ be the Standard-Decoding distribution and $q_{\rm Edit}(y\mid c,f)$ the editing distribution supplied with future context $f$. Define
\begin{align}
  \mathcal L_{\rm Std}
  &=\mathbb E[-\log q_{\rm Std}(Y_i\mid C_i)],\notag\\
  \mathcal L_{\rm Oracle}
  &=\mathbb E[-\log q_{\rm Edit}(Y_i\mid C_i,F_i^\star)],\notag\\
  \mathcal L_{\rm MBE}
  &=\mathbb E[-\log q_{\rm Edit}(Y_i\mid C_i,\widetilde F_i)].
  \label{eq:three_losses}
\end{align}
The conditional modeling errors under standard and oracle-context editing are
\begin{align}
  \epsilon_{\rm Std}
  &=\mathbb E_{C_i}D_{\rm KL}\!\left(
    p(\cdot\mid C_i)\,\|\,q_{\rm Std}(\cdot\mid C_i)\right),\notag\\
  \epsilon_{\rm Oracle}
  &=\mathbb E_{C_i,F_i^\star}D_{\rm KL}\!\left(
    p(\cdot\mid C_i,F_i^\star)\,\|\,
    q_{\rm Edit}(\cdot\mid C_i,F_i^\star)\right).
  \label{eq:model_mismatch}
\end{align}
Define the predicted-context term
\begin{equation}
  \delta_{\rm pred}:=\mathcal L_{\rm MBE}-\mathcal L_{\rm Oracle}.
  \label{eq:prediction_penalty}
\end{equation}
It measures the change in editing loss when ground-truth future context is replaced by its generated approximation; positive values constitute a prediction penalty. Cross-entropy decomposition gives
\begin{equation}
  \boxed{\begin{aligned}
  \mathcal L_{\rm Std}-\mathcal L_{\rm MBE}
  &=I(Y_i;F_i^\star\mid C_i)\\
  &\quad-\left(\epsilon_{\rm Oracle}-\epsilon_{\rm Std}\right)
  -\delta_{\rm pred}.
  \end{aligned}}
  \label{eq:practical_gain}
\end{equation}
The terms respectively capture oracle information gain, modeling mismatch under the wider editing pattern, and replacement of ground-truth context by generated tokens.

\noindent\textbf{Proposition 2 (Benefit under sufficiently accurate future prediction).}
If the editing loss is $L_F$-Lipschitz with respect to a future-context distance $d$, then
\begin{equation}
  \delta_{\rm pred}\leq L_F\mathbb E[d(\widetilde F_i,F_i^\star)],
  \label{eq:prediction_penalty_bound}
\end{equation}
Combining this bound with Equation~\eqref{eq:practical_gain}, MBE has lower expected log loss whenever
\begin{equation}
  L_F\mathbb E[d(\widetilde F_i,F_i^\star)]
  +\epsilon_{\rm Oracle}-\epsilon_{\rm Std}
  <I(Y_i;F_i^\star\mid C_i).
  \label{eq:mbe_benefit_condition}
\end{equation}
The generated future need not be exact: its induced editing error and the cross-block modeling mismatch must together remain below the oracle value of missing future information. The proof and the resulting lower bound are in the Supplementary Material.

\noindent\textbf{Connection to MBE design.}
Theorem~1 establishes the potential value of later context, while Proposition~2 separates the practical effects of generated-context quality and cross-block modeling mismatch. MBE supplies $\widetilde F_i$ through cross-block editing, and Multi-Block Edit SFT aligns the model with the wider attention pattern.

\begin{table*}[!t]
\centering
\small
\setlength{\tabcolsep}{2.0pt}
\begin{tabular}{@{}l*{13}{c}@{}}
\toprule
& \multicolumn{3}{c}{\textbf{Reasoning}} & \multicolumn{4}{c}{\textbf{Coding}} &
\multicolumn{3}{c}{\textbf{Math}} & \multicolumn{2}{c}{\textbf{Agent}} & \\
\cmidrule(lr){2-4}\cmidrule(lr){5-8}\cmidrule(lr){9-11}\cmidrule(lr){12-13}
\textbf{Method} & \textbf{Zebra} & \textbf{OCNLI} & \textbf{DROP} & \textbf{MBPP+} &
\textbf{HE+} & \textbf{LCB v6} & \textbf{Spider} & \textbf{AIME} & \textbf{GSM+} &
\textbf{Omni} & \textbf{BFCL} & \textbf{Nexus} & \textbf{Avg.}\\
\midrule
Standard Decoding & 68.50 & 61.02 & 81.55 & 73.28 & 80.49 & 28.85 & 75.78 & \underline{36.67} & 85.88 & 41.70 & 72.06 & 31.59 & 61.45\\
ReMDM & \underline{72.40} & 61.08 & 81.75 & 72.75 & 80.49 & 29.30 & \underline{76.71} & 33.33 & \textbf{86.23} & \underline{42.60} & \underline{72.50} & \textbf{32.15} & 61.77\\
SABER & 70.20 & 61.22 & \textbf{82.11} & 71.16 & \underline{82.93} & 29.07 & 76.01 & 33.33 & 85.95 & 42.20 & 72.19 & \underline{32.14} & 61.54\\
DCD & 64.60 & 60.44 & 81.21 & 71.43 & 79.88 & 27.75 & 75.64 & 26.67 & 85.60 & 40.70 & 72.23 & 32.11 & 59.86\\
\midrule
\textbf{Training-Free MBE} & 70.40 & \underline{61.86} & \underline{81.96} & \underline{74.34} & \underline{82.93} & \underline{29.74} & 76.67 & \underline{36.67} & \underline{86.04} & 42.10 & \textbf{73.08} & 31.88 & \underline{62.31}\\
\textbf{MBE} & \textbf{74.40} & \textbf{62.71} & 80.93 & \textbf{75.40} & \textbf{84.76} & \textbf{31.06} & \textbf{77.69} & \textbf{50.00} & 85.75 & \textbf{44.20} & 71.88 & 32.09 & \textbf{64.24}\\
\bottomrule
\end{tabular}
\caption{Overall performance on LLaDA2.1-Mini. Baselines appear above the rule and MBE variants below; only MBE uses the post-trained checkpoint. Bold and underlining mark the best and second-best results.}
\label{tab:main}
\end{table*}

\section{Experiments}
\label{sec:experiments}

\subsection{Experimental Setup}

\paragraph{Models and decoding.}
All methods use zero-shot prompting with one generation per example ($n{=}1$, pass@1), deterministic temperature-zero decoding, and a maximum generation length of 32,768 tokens. We primarily evaluate LLaDA2.1-Mini~\citep{bie2026llada2} with block size $B{=}32$, and use LLaDA2.1-Flash for a model-scale check. All methods share $\omega_{\rm mask}{=}0.5$ and $\omega_{\rm edit}{=}0.0$. Standard Decoding on either checkpoint retains up to 16 intra-block T2T-only post-edit passes after mask resolution. Training-Free MBE and MBE disable that phase and instead use at most $N_{\rm MBE}{=}5$ cross-block passes. Training-Free MBE uses the base checkpoint and a fixed $K{=}2$ window.

\paragraph{Post-training.}
Multi-Block Edit SFT is trained for six epochs on an 8-million-sample subset of the original LLaDA2.1 SFT corpus, excluding samples from every benchmark evaluated in this paper. Its three two-epoch phases use $(K,\alpha)=(2,0.05)$, $(3,0.15)$, and $(\{3,4\},0.25)$. MBE uses this checkpoint with $K_{\max}{=}4$ and tail-entropy block selection at $\rho{=}0.25$; we also evaluate the same checkpoint under Standard Decoding to isolate post-training gains from cross-block editing.

\paragraph{Baselines and benchmarks.}
We compare with Standard Decoding, ReMDM~\citep{schiff2025remdm}, SABER~\citep{xiao2025saber}, and DCD~\citep{gong2025dcd}. The 12 benchmarks cover reasoning (ZebraLogic~\citep{zebralogic}, OCNLI~\citep{ocnli}, DROP~\citep{drop}), code and structured generation (MBPP+ and HumanEval+~\citep{liu2023humameval+mbpp}, LiveCodeBench v6~\citep{livecodebench}, Spider~\citep{yu2018spider}), mathematics (AIME~2025~\citep{aime2025}, GSM-Plus~\citep{li2024gsm}, Omni-MATH~\citep{omnimath}), and function calling (BFCL~v3~\citep{bfcl}, Nexus FCB~\citep{nexusfcb}).

\subsection{Main Results}

Tables~\ref{tab:main} and~\ref{tab:matched_control} compare methods across 12 benchmarks with matched base and post-trained checkpoints. Training-Free MBE tests whether one generated block can support revision at $K{=}2$; Standard Decoding on the MBE-SFT checkpoint isolates post-training from cross-block editing. Each matched comparison improves 11 benchmarks and ties one, showing a consistent task-level direction.

\paragraph{Training-Free MBE improves consistently.}
Training-Free MBE raises the base-checkpoint average from 61.45 to 62.31 and improves or preserves all 12 benchmarks. With weights fixed, these gains place two-block editing in the favorable regime of Section~\ref{sec:theory}, where the correction value of the next block exceeds prediction error and attention shift. Larger gains on HumanEval+, ZebraLogic, MBPP+, and BFCL are consistent with later spans exposing earlier syntactic or constraint violations.

\paragraph{Post-training extends the benefit of MBE.}
The complete framework reaches 64.24. On the same checkpoint, MBE adds 2.14 points over Standard Decoding (62.10), whereas post-training alone adds 0.65. Their interaction indicates that Multi-Block Edit SFT chiefly teaches the model to use wider context for correction rather than acting as generic task fine-tuning. Although the post-trained standard baseline is weaker on DROP, GSM-Plus, and BFCL, MBE improves every matched comparison.

\begin{table*}[!t]
\centering
\small
\setlength{\tabcolsep}{1.8pt}
\begin{tabular}{@{}ll*{13}{c}@{}}
\toprule
& & \multicolumn{3}{c}{\textbf{Reasoning}} & \multicolumn{4}{c}{\textbf{Coding}} &
\multicolumn{3}{c}{\textbf{Math}} & \multicolumn{2}{c}{\textbf{Agent}} & \\
\cmidrule(lr){3-5}\cmidrule(lr){6-9}\cmidrule(lr){10-12}\cmidrule(lr){13-14}
\textbf{Checkpoint} & \textbf{Decoder} & \textbf{Zebra} & \textbf{OCNLI} & \textbf{DROP} & \textbf{MBPP+} &
\textbf{HE+} & \textbf{LCB v6} & \textbf{Spider} & \textbf{AIME} & \textbf{GSM+} &
\textbf{Omni} & \textbf{BFCL} & \textbf{Nexus} & \textbf{Avg.}\\
\midrule
LLaDA2.1-Mini & Standard
& 68.50
& 61.02
& \textbf{81.55}
& 73.28
& 80.49
& 28.85
& 75.78
& 36.67
& \textbf{85.88}
& 41.70
& \textbf{72.06}
& 31.59
& 61.45\\

MBE-SFT & Standard
& 71.30
& \textbf{62.71}
& 80.31
& 71.96
& 84.15
& 27.09
& 76.76
& 40.00
& 85.04
& 43.10
& 71.57
& 31.21
& 62.10\\

MBE-SFT & MBE
& \textbf{74.40}
& \textbf{62.71}
& 80.93
& \textbf{75.40}
& \textbf{84.76}
& \textbf{31.06}
& \textbf{77.69}
& \textbf{50.00}
& 85.75
& \textbf{44.20}
& 71.88
& \textbf{32.09}
& \textbf{64.24}\\
\bottomrule
\end{tabular}
\caption{Decoder--checkpoint ablation. The two MBE-SFT rows share the post-trained checkpoint and differ only in decoding.}
\label{tab:matched_control}
\end{table*}

\subsection{Generalization}

\begin{table*}[!t]
\centering

\begin{minipage}[t]{0.35\textwidth}
\vspace{0pt}
\centering

{%
\footnotesize
\setlength{\tabcolsep}{2.5pt}
\renewcommand{\arraystretch}{1.05}

\begin{tabular}{@{}crrr@{\hspace{6pt}}rrr@{}}
\toprule
& \multicolumn{3}{c}{\textbf{HumanEval+}}
& \multicolumn{3}{c}{\textbf{MBPP+}}\\
\cmidrule(lr){2-4}
\cmidrule(lr){5-7}
\textbf{$B$}
& \textbf{Std.}
& \textbf{MBE}
& \textbf{$\Delta$}
& \textbf{Std.}
& \textbf{MBE}
& \textbf{$\Delta$}\\
\midrule
8  & 78.05 & 79.27 & +1.22
   & 71.16 & 73.81 & +2.65\\
16 & 75.61 & 83.54 & +7.93
   & 71.96 & 74.87 & +2.91\\
32 & 80.49 & 82.93 & +2.44
   & 73.28 & 74.34 & +1.06\\
\bottomrule
\end{tabular}
}

\caption{MBE across native block sizes on the base
checkpoint.}
\label{tab:block_size}
\end{minipage}
\hspace{0.025\textwidth}
%
\begin{minipage}[t]{0.255\textwidth}
\vspace{0pt}
\centering

{%
\footnotesize
\setlength{\tabcolsep}{2.7pt}
\renewcommand{\arraystretch}{1.05}

\begin{tabular}{@{}lrrr@{}}
\toprule
\textbf{Benchmark}
& \textbf{Std.}
& \textbf{MBE}
& \textbf{$\Delta$}\\
\midrule
ZebraLogic & 84.20 & 88.20 & +4.00\\
MBPP+      & 76.72 & 79.10 & +2.38\\
GSM-Plus   & 89.23 & 89.61 & +0.38\\
Nexus FCB  & 44.83 & 46.62 & +1.79\\
\bottomrule
\end{tabular}
}

\caption{MBE on the LLaDA2.1-Flash base checkpoint.}
\label{tab:flash}
\end{minipage}
\hspace{0.025\textwidth}
%
\begin{minipage}[t]{0.285\textwidth}
\vspace{0pt}
\centering

{%
\footnotesize
\setlength{\tabcolsep}{3.0pt}
\renewcommand{\arraystretch}{1.05}

\begin{tabular}{@{}lrr@{}}
\toprule
\textbf{Training variant}
& \textbf{HE+}
& \textbf{MBPP+}\\
\midrule
Standard block SFT
& 81.71 & 74.34\\
+ attention mask
& 82.93 & 74.34\\
+ correction bonus
& \textbf{85.37} & 74.57\\
+ window curriculum
& 84.76 & \textbf{75.40}\\
\bottomrule
\end{tabular}
}

\caption{Cumulative ablation of MBE post-training
components.}
\label{tab:mbe_sft_ablation}
\end{minipage}

\end{table*}


\paragraph{Block size.}
All six matched comparisons in Table~\ref{tab:block_size} favor MBE at block sizes 8, 16, and 32. Gains are non-monotonic because $B$ jointly changes the information crossing a boundary, the next block's accuracy, and the Standard-Decoding baseline; nevertheless, the information--error balance favors MBE throughout.


\paragraph{Model scale.}
MBE improves all four evaluated tasks on the stronger LLaDA2.1-Flash checkpoint (Table~\ref{tab:flash}), extending the benefit of cross-block correction to a larger model.


\begin{table}[!t]
\centering
\small
\setlength{\tabcolsep}{2.2pt}
\begin{tabular}{@{}crr@{\hspace{8pt}}crr@{}}
\toprule
\multicolumn{3}{c}{\textbf{(a) Editing window}} &
\multicolumn{3}{c}{\textbf{(b) Pass budget}}\\
\cmidrule(lr){1-3}\cmidrule(lr){4-6}
\textbf{$K$} & \textbf{HE+} & \textbf{MBPP+} &
\textbf{$N_{\rm MBE}$} & \textbf{HE+} & \textbf{MBPP+}\\
\midrule
1 & 80.49 & 73.28 & 1 & 81.10 & 72.75\\
2 & \textbf{82.93} & \textbf{74.34} & 3 & \textbf{82.93} & 73.81\\
3 & 79.26 & 73.55 & 5 & \textbf{82.93} & \textbf{74.34}\\
4 & 76.83 & 72.88 & 7 & \textbf{82.93} & \textbf{74.34}\\
\bottomrule
\end{tabular}

\vspace{3pt}
\begin{tabular}{@{}lrrrr@{}}
\toprule
\multicolumn{5}{c}{\textbf{(c) Window policy}}\\
\cmidrule(lr){1-5}
\textbf{Strategy} & \textbf{HE+} & \textbf{MBPP+} & \textbf{AIME} & \textbf{Avg.}\\
\midrule
Random span & 82.32 & 74.07 & 40.00 & 65.46\\
Fixed 4-block span & 83.54 & 75.13 & 40.00 & 66.22\\
Tail-entropy selection & \textbf{84.76} & \textbf{75.40} & \textbf{50.00} & \textbf{70.05}\\
\bottomrule
\end{tabular}
\caption{Editing-policy ablations. Panels (a,b) use the base checkpoint (panel (b): $K{=}2$); panel (c) uses the MBE-SFT checkpoint.}
\label{tab:refinement_ablation}
\end{table}

\begin{table}[!t]
\centering
\small
\setlength{\tabcolsep}{3.2pt}
\begin{tabular}{@{}lrrr@{}}
\toprule
\multicolumn{4}{c}{\textbf{(a) End-to-end throughput by dataset}}\\
\cmidrule(lr){1-4}
\textbf{Dataset} & \textbf{Std. TPS} & \textbf{MBE TPS} & \textbf{Retention}\\
\midrule
GSM-Plus & 887.54 & 842.03 & 94.9\%\\
HumanEval+ & 1303.86 & 1138.86 & 87.3\%\\
AIME~2025 & 1060.81 & 1001.91 & 94.4\%\\
LiveCodeBench v6 & 882.00 & 852.62 & 96.7\%\\
\bottomrule
\end{tabular}

\vspace{3pt}
\begin{tabular}{@{}lrr@{}}
\toprule
\multicolumn{3}{c}{\textbf{(b) Runtime components on HumanEval+}}\\
\cmidrule(lr){1-3}
\textbf{Runtime variant} & \textbf{TPS} & \textbf{Retention}\\
\midrule
No optimization & 590.16 & 45.3\%\\
Cache-safe revision only & 690.35 & 52.9\%\\
Multi-shape graphs only & 893.25 & 68.5\%\\
Full MBE runtime & 1138.86 & 87.3\%\\
\bottomrule
\end{tabular}
\caption{End-to-end throughput and runtime ablation. Panel (a) includes all overhead; panel (b) evaluates optimizations separately and jointly.}
\label{tab:dataset_tps_main}
\end{table}

\begin{figure}[!t]
  \centering
  \includegraphics[width=0.90\columnwidth]{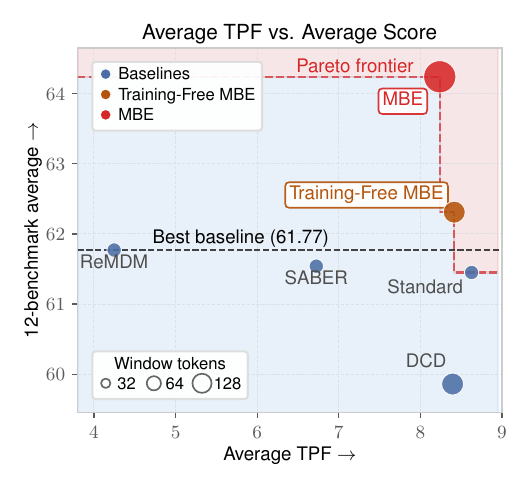}
  \caption{Quality--TPF trade-off. Accuracy averages Table~\ref{tab:main}; TPF averages BFCL~v3, HumanEval+, GSM-Plus, and Spider. Marker area denotes maximum window length.}
  \label{fig:tpf_score}
\end{figure}

\subsection{Ablation Studies}

We ablate cross-block training alignment, editing-window allocation, and refinement convergence.

\paragraph{Training alignment.}
With data and decoding fixed, each cumulative component in Table~\ref{tab:mbe_sft_ablation} raises the two-benchmark average. The attention mask reduces cross-block mismatch, the correction bonus emphasizes future-exposed errors, and the window curriculum aligns adaptive editing spans.

\paragraph{Window length.}
With the base checkpoint fixed, panel (a) shows that $K{=}2$ is optimal. Equation~\eqref{eq:practical_gain} explains why wider windows need not help the unadapted model: potentially useful context competes with error from longer generated spans and a less familiar attention pattern, motivating post-training for broader editing.

\paragraph{Refinement budget.}
Panel (b) improves through three to five passes and is unchanged at seven, indicating that the edit loop has converged within the adopted five-pass budget.

\paragraph{Block selection.}
Panel (c) asks how far revision should reach. A fixed four-block span outperforming random spans favors contiguous editing context. Tail-entropy selection retains that continuity but adapts its starting block; its further gains indicate that revision demand varies across committed blocks and that uncertainty usefully allocates the editing horizon.

\subsection{Efficiency}
\label{sec:efficiency}


\paragraph{Tokens per forward (TPF).}
Figure~\ref{fig:tpf_score} compares algorithmic efficiency. Both MBE variants lie on the quality--TPF frontier; MBE also exceeds Standard Decoding on BFCL~v3 and Spider as fewer later-block forwards offset its edit passes.

\paragraph{Tokens per second (TPS).}
Table~\ref{tab:dataset_tps_main}(a) shows that MBE retains 87.3--96.7\% of Standard-Decoding TPS. Panel (b) attributes the remaining overhead: cache-safe revision avoids rebuilding unchanged history, while multi-shape graphs keep editing forwards on the CUDA Graph path. Together they raise retention from 45.3\% to 87.3\% on HumanEval+.

\section{Related Work}

\paragraph{Diffusion decoding and refinement.}
Masked dLLMs denoise tokens in parallel~\citep{nie2025large,ye2025dream}, while block diffusion bounds generation and LLaDA2.1 adds M2T/T2T editing~\citep{arriola2025block,bie2026llada2}. ReMDM, SABER, and Hierarchy Decoding alter parallel refinement~\citep{schiff2025remdm,xiao2025saber,qi2026hierarchy}; MBE revises completed blocks using later context.

\paragraph{Commitment and future context.}
DCD delays commitment, whereas VSB selects self-contained boundaries~\citep{gong2025dcd,wang2026vsb}; both act before commitment. MBE instead revises fixed, completed blocks using later context, supported by cross-block training and cache-safe execution. ParallelBench studies ignored-dependency losses~\citep{kang2026parallelbenchunderstandingtradeoffsparallel}. Extended comparisons are in the Supplement.

\section{Conclusion}

Fixed-block decoding is efficient, yet future-context sensitivity and uncertainty concentrate near block ends. MBE uses later context to reopen completed blocks, aligns cross-block revision through post-training, and runs with a cache-safe runtime. Across 12 benchmarks, it gains 2.79 points while retaining 87.3--96.7\% of Standard-Decoding throughput, improving quality without forfeiting blockwise efficiency.

\FloatBarrier
\bibliography{references}
\end{document}